\documentclass[conference]{IEEEtran}
\IEEEoverridecommandlockouts

\usepackage{cite}
\usepackage{amsmath,amssymb,amsfonts}
\usepackage{algorithmic}
\usepackage{graphicx}
\usepackage{textcomp}
\usepackage{xcolor}
\usepackage{listings}
\usepackage{url}
\usepackage{booktabs}
\usepackage{microtype}          
\usepackage[htt]{hyphenat}      
\usepackage[hidelinks]{hyperref}

\graphicspath{{report/figures/}{./}}

\begin{document}

\title{Correct First, Fast Second:\\
What a Local 7B Code Model Can and Cannot Do\\
for GPU Kernel Optimization}

\author{%
\IEEEauthorblockN{Jui-Chien Tsou \quad Po-Yi Wu}
\IEEEauthorblockA{Department of Computer Science and Information Engineering\\
National Taiwan University\\
\{b11902044, b11902127\}@csie.ntu.edu.tw}%
}
\maketitle

\begin{abstract}
Most published results on LLM-generated GPU kernels come from frontier-scale
models or from models fine-tuned with reinforcement learning. We ask a different
question: if the model is a fixed, locally hosted 7B code model
(Qwen2.5-Coder-7B-Instruct), how much does the way we drive it change what it can
do? On 30 KernelBench tasks (the first 10 of levels 1--3, V100, fp32) we compare
four approaches with increasing structure: zero-shot prompting, prompting with
profiler output, iterative refinement on execution feedback, and a multi-agent
loop in which the same model plays four roles (analysis, retrieval, generation,
feedback triage) around a compile-and-run evaluator. We find, first, that
correctness rather than speed is the limiting factor. Even correct kernels
average 0.97--0.99$\times$ the speed of eager PyTorch, because a 7B model cannot
beat cuBLAS/cuDNN on dense GEMM or convolution; the only speedups come from
fusing memory-bound epilogues. Second, the best approach depends on task
difficulty. Profiler-guided prompting is strongest overall (43\% correct, 27\%
faster than baseline), but the agent loop is the only method that solves any
level-2 fusion task (3/10; every other method 0/10). Third, a naively built agent
is worse than no agent. Our first agent scored 3\% correct, below zero-shot's
30\%. What fixed it was matching the agent's ambition to the model's ability and
spending compute on best-of-$n$ sampling judged by the real evaluator
(3\%$\to$33\% correct from $n{=}1$ to $n{=}8$), rather than on more reasoning
steps. We release code, prompts, generations, and evaluation logs.
\end{abstract}


\section{Introduction}

Writing a fast GPU kernel is expert work. A CUDA kernel has to get memory
coalescing, shared-memory capacity, occupancy, and numerical semantics right at
the same time, and it has to beat the vendor libraries (cuBLAS,
cuDNN~\cite{cudnn}) that PyTorch~\cite{pytorch} already calls. Language models
may be able to do some of this work, and KernelBench~\cite{kernelbench} turns it
into a benchmark: given a PyTorch reference module \texttt{Model}, write a
\texttt{ModelNew} that computes the same function with custom CUDA and runs
faster.

Most reported progress comes from frontier proprietary models driven by expensive
evolutionary or agentic search~\cite{aicudaengineer,astra}, or from models
post-trained with reinforcement learning on KernelBench-style tasks, such as
Kevin's 32B model~\cite{kevin} and CUDA-L1~\cite{cudal1}. Someone with one
workstation, a confidentiality requirement, or a fixed inference budget is in a
different position. Their model is small and open-weight, and they are not going
to retrain it. What they can change is the scaffold: what the model is told, how
many attempts it gets, and who judges the attempts. So we ask: with one mid-size
open-weight model held fixed, how much do prompting and agent scaffolding matter,
and where do they help or hurt?

We hold the model fixed and compare four conditions on the same 30 tasks:
zero-shot prompting, profiler-guided prompting, iterative refinement on execution
feedback, and a multi-agent loop in which the same 7B model plays four roles (code
analyzer, retrieval researcher, kernel generator, feedback triager) around a real
compile-and-run evaluator. An agent's instructions are a design choice too, so we
also report how ours came about: a first version that scored below zero-shot, and
the redesign that fixed it.

Our contributions are:
\begin{enumerate}
  \item a controlled $1{\times}4$ comparison on 30 KernelBench tasks with compile,
        correctness, and speedup metrics, a per-difficulty breakdown, and the
        compute cost of each method;
  \item evidence that no single method wins everywhere: profiler-guided prompting
        does best overall and on whole networks, while only the agent solves fused
        multi-operator tasks;
  \item an analysis of why a naively built agent scores below zero-shot, and a
        repair (competence-matched prompting plus best-of-$n$ sampling under an
        execution oracle) checked with a one-component-at-a-time ablation and a
        test-time-compute scaling curve;
  \item a practical warning: LLM-generated kernels can deadlock the GPU, so
        unattended evaluation needs a hard per-candidate timeout.
\end{enumerate}

\section{Background and Related Work}

\subsection{Problem setting}

\textbf{Benchmark.} KernelBench~\cite{kernelbench} contains 250 PyTorch reference
modules in three difficulty levels: level~1 has single operators (GEMM variants,
convolutions, elementwise ops), level~2 has short fused operator sequences (e.g.\
\texttt{ConvTranspose2d}$\to$\allowbreak\texttt{BiasAdd}$\to$\allowbreak\texttt{Clamp}$\to$\allowbreak\texttt{Scale}),
and level~3 has full architectures (MLPs, LeNet-5, AlexNet, \dots). Each task
ships \texttt{get\_inputs()} generators; a candidate passes if its outputs match
the reference within tolerance across randomized trials. The headline metric is
\emph{fast$_p$}, the fraction of tasks that are both correct and faster than a
baseline by a factor $p$; we use $p{=}1$ against eager PyTorch. We use the
\textbf{first 10 problems of each of levels 1--3}, 30 tasks in all. The subset
keeps the full study (four methods plus eight agent configurations for the
ablations and the best-of-$n$ sweep) affordable on shared GPUs while still
covering three kinds of task.

\textbf{Baseline.} The performance baseline is eager PyTorch, whose matmul and
convolution calls go to highly tuned cuBLAS/cuDNN kernels~\cite{cudnn}. For
compute-bound operators this baseline is out of reach for a 7B model, as we show
below; the roofline model~\cite{roofline} says custom code can only win where the
reference is memory-bound or launch-bound.

\textbf{Model and tooling.} All generation uses
Qwen2.5-Coder-7B-Instruct~\cite{qwen25coder} served locally through HuggingFace
Transformers (fp16, multi-GPU workers; one GPU generates while a second
evaluates). No proprietary model or API is used anywhere. The agent framework is
about 1{,}500 lines of Python that orchestrate the roles described in
\S\ref{sec:method}; retrieval uses BM25~\cite{bm25} over a small curated corpus of
CUDA optimization notes (tiling, coalescing, epilogue fusion,
\texttt{torch.utils.cpp\_extension} usage).

\subsection{Related work}

\textbf{LLMs for GPU kernel generation.} KernelBench~\cite{kernelbench} showed
that even frontier reasoning models match the PyTorch baseline on fewer than 20\%
of tasks out of the box, and that execution and profiling feedback help. Sakana's
AI CUDA Engineer~\cite{aicudaengineer} drives a frontier LLM with evolutionary
search and a kernel archive; the same group later introduced
robust-kbench~\cite{robustkbench} after finding that existing benchmark harnesses
admitted exploitable loopholes. Astra~\cite{astra} optimizes production SGLang
kernels with a multi-agent system around OpenAI o4-mini. On the training side,
Kevin~\cite{kevin} applies multi-turn RL to QwQ-32B and CUDA-L1~\cite{cudal1}
applies contrastive RL with speedup rewards; for the Triton DSL,
TritonBench~\cite{tritonbench} provides a benchmark, AutoTriton~\cite{autotriton}
trains an 8B model with RL, and AMD's GEAK~\cite{geak} uses a Reflexion-style
agent with inference-time scaling. CUDA-LLM~\cite{cudallm} combines feature
search with compile, correctness, and latency feedback on DeepSeek-V3. Every one
of these either uses a frontier model or trains the model. We hold an untrained
7B model fixed and vary only the scaffold, which shows what scaffolding alone
does.

\textbf{Test-time compute and verifiers.} Sampling many candidates and selecting
with a verifier is well established for code: pass@$k$ scaling in
Codex~\cite{codex}, filtering and clustering with generated tests in
AlphaCode~\cite{alphacode} and CodeT~\cite{codet}, and the coverage curves of
Large Language Monkeys~\cite{monkeys}. Snell et al.~\cite{snell2024} show that
the best way to spend test-time compute depends on problem difficulty and on the
base model. We see the same in a harder setting: with a weak generator and a
compile-and-run verifier, more samples beat more reasoning, and the best strategy
changes with difficulty.

\textbf{Iterative and agentic code generation.} Self-Refine~\cite{selfrefine},
Self-Debug~\cite{selfdebug}, and Reflexion~\cite{reflexion} refine code from
feedback; MetaGPT~\cite{metagpt} and related frameworks assign specialized roles
to cooperating LLM agents; retrieval-augmented generation~\cite{rag} grounds
generation in documents. Our agent uses all three, and the ablation in
\S\ref{sec:ablation} measures what each one is worth when the model is small.

\textbf{Classical kernel optimization.} Hand-tuned libraries~\cite{cudnn},
tile-level DSLs such as Triton~\cite{triton}, and auto-tuning compilers such as
TVM~\cite{tvm} and Ansor~\cite{ansor} are the established routes to fast kernels.
Hand-written kernels beat libraries most reliably on memory-bound fusion,
FlashAttention~\cite{flashattention} being the best-known case, and that is also
the only place our small model wins.

\section{Methodology}\label{sec:method}

All four methods share the same base task prompt (Appendix~\ref{app:prompt}),
model, decoding budget (4096 new tokens), and evaluator. They differ only in what
the model is told and in the control flow around it.

\subsection{The four interaction regimes}

\textbf{Zero-shot.} The raw KernelBench prompt, greedy decoding, one attempt.
This is the floor.

\textbf{Guided.} We profile each reference module with the PyTorch profiler on
the target GPU and add its top operators by CUDA time to the prompt, followed by
a roofline-style rule~\cite{roofline} we wrote after the pilot study
(\S\ref{sec:iterate}): do not reimplement matmul or conv, since cuBLAS/cuDNN will
win; fuse the cheap memory-bound epilogue instead. One attempt, no loop.

\textbf{Iterative.} Up to three turns: generate, compile/check/time on the GPU,
feed a condensed error and timing report back, regenerate. The submitted kernel is
the best turn under a $(\textit{correct},\textit{compiled},\textit{speedup})$
ranking.

\begin{figure*}[t]
  \centering
  \includegraphics[width=\textwidth]{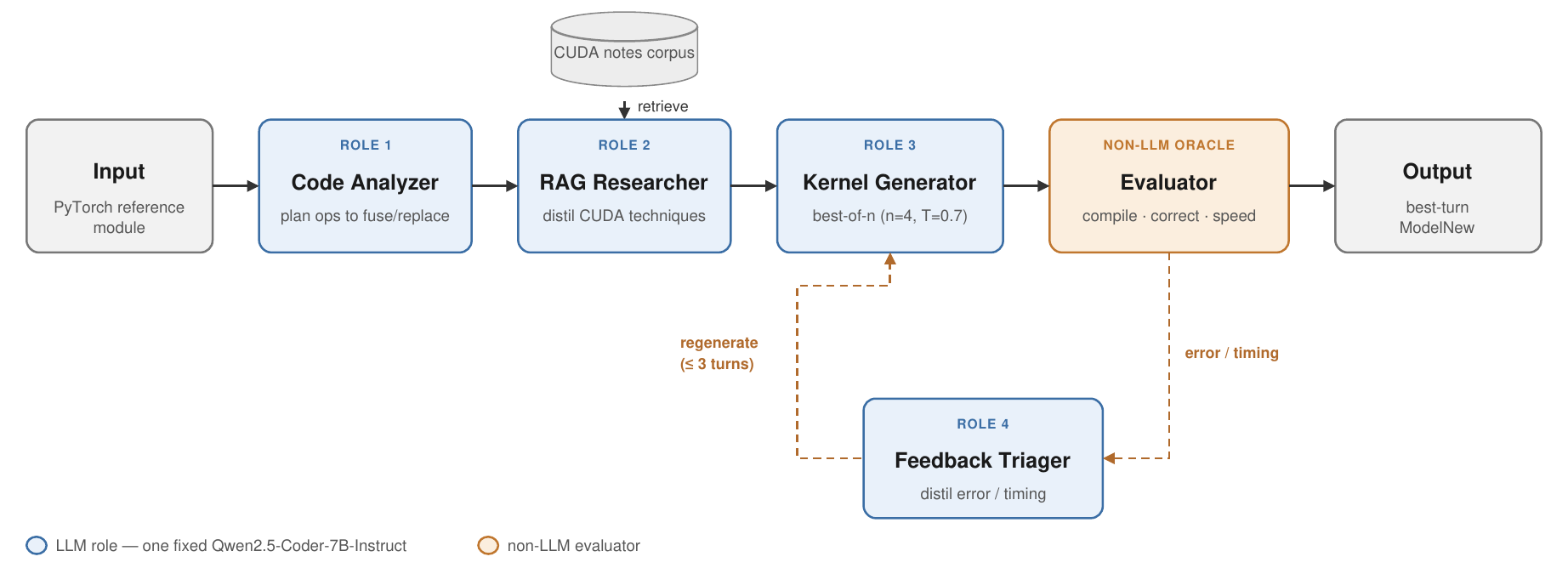}
  \caption{The agentic loop. Four roles are the same Qwen2.5-Coder-7B-Instruct under
  different system prompts (blue); a non-LLM Evaluator (orange) compiles, checks, and
  times each candidate and closes the feedback loop. Generation is best-of-$n$
  ($n{=}4$); the loop runs at most three turns.}
  \label{fig:pipeline}
\end{figure*}

\textbf{Agentic.} A five-stage loop: four roles played by the same model under
different system prompts, plus a non-LLM Evaluator that closes the loop
(Fig.~\ref{fig:pipeline}; Appendix~\ref{app:sysprompts}):
\begin{enumerate}
  \item \emph{Code Analyzer} -- names the operations, bottlenecks, and fusion
        opportunities; emits retrieval queries.
  \item \emph{RAG Researcher} -- retrieves from the documentation corpus and writes
        a cited optimization brief.
  \item \emph{Kernel Generator} -- writes \texttt{ModelNew}; samples $n$ candidates
        per turn (best-of-$n$).
  \item \emph{Evaluator} (non-LLM) -- compiles, checks correctness, and times each
        candidate; keeps the best per turn.
  \item \emph{Feedback Analyzer} -- converts raw evaluator output into a diagnosis
        and a directive for the next turn.
\end{enumerate}
The loop runs at most three turns. Each component can be switched off on its own,
which is how the ablation in \S\ref{sec:ablation} is built.

\subsection{How the agent was developed}\label{sec:iterate}

Readers of any agent paper should want to know where its instructions came from.
Ours went through one major revision: a first agent that failed badly, then a
redesign (full chronology in Appendix~\ref{app:mods}). We report both because
the failure tells us as much as the fix.

\textbf{Pilot (v1): 3\% correct, 33\% compile, below zero-shot.} The v1 generator
prompt required custom kernels (``replace PyTorch operators \dots using
\texttt{load\_inline} or Triton''), and the analyzer and researcher briefs
recommended tiling, shared memory, and warp reductions on every task. The 7B
model then attempted hand-tiled CUDA it could not compile: it called
\texttt{load\_inline(sources=...)}, a keyword that does not exist, and mismatched
\texttt{PYBIND11\_MODULE} names, which produced \texttt{PyInit} import errors.
The feedback loop could not recover because each retry made the same kind of
attempt. A single-turn variant actually beat the full loop.

\textbf{Repairs (v2).} We made four changes, each in response to a specific
failure. (1)~\emph{Competence-matched prompting}: every role is told to reach a
correct, compiling kernel first (reuse \texttt{torch} ops for the heavy compute,
fuse only the memory-bound epilogue) and to move to hand-written tiling only once
a correct baseline exists; the prompts also spell out the exact
\texttt{load\_inline} signature the model kept getting wrong.
(2)~\emph{Best-of-$n$ sampling}: $n{=}4$ candidates per turn at temperature~0.7,
ranked by the real evaluator. (3)~\emph{Monotone turn selection}: the submitted
kernel is the best turn by $(\textit{correct},\textit{compiled},\textit{speedup})$,
so a later broken retry can never replace an earlier working kernel. This bug had
also been quietly hurting the iterative baseline, and we fixed it there too.
(4)~\emph{Hard evaluation timeouts}: each candidate runs in a child process that
is killed after 240\,s, because generated kernels can deadlock the GPU. One such
kernel stalled a production run for 14 hours (\S\ref{sec:infra}).

\section{Experimental Setup}

\textbf{Hardware/software.} Generation and evaluation run on nodes with 8$\times$
NVIDIA V100-SXM2-32GB (Volta, no fp32 tensor cores), CUDA 12.3, PyTorch 2.x, GCC
10. Workers pair GPUs: one serves the model and one evaluates kernels, so
generation never competes with timing.

\textbf{Measurement.} Each candidate is compiled with PyTorch's
\texttt{cpp\_extension} module, checked for correctness over 5 trials with
randomized inputs (outputs must match the reference within KernelBench's default
tolerance), and timed with CUDA events over 10 trials after warm-up, in fp32. The
speed baseline is the eager-PyTorch reference time measured on the same GPU.

\textbf{Success criteria.} We report compile rate; correctness rate;
\emph{fast$_1$}, meaning correct \textbf{and} faster than eager PyTorch, which is
the success definition for this study; and geometric-mean speedup over correct
samples. A task counts once per (method, task) cell. Multi-turn and multi-sample
methods submit exactly one final kernel, chosen by the evaluator and never by the
LLM's own judgment.

\textbf{Compute budget.} The four methods are not compute-matched, by design.
Zero-shot and guided make one generator call per task (guided also runs the
profiler once on the reference); iterative makes at most three; the agent makes
at most $3\times4{=}12$ generator calls plus the analyzer, researcher, and
feedback-analyzer calls, and evaluates every sampled candidate.
Table~\ref{tab:headline} lists the generator budget next to accuracy. So the
agent-versus-prompting comparison answers ``what does the extra structure and
compute get you,'' and the best-of-$n$ sweep (Fig.~\ref{fig:bestof}) is the
compute-scaling curve.

\textbf{Reproducibility.} One command reruns the full matrix; the report, tables,
and figures are regenerated from raw evaluation JSON by a single script. Completed
cells are skipped, so the study can be re-run incrementally. Code and artifacts
are released (see Code and Data Availability).

\section{Results}

\subsection{Headline comparison}

\begin{table}[t]
\caption{Four methods on 30 KernelBench tasks (levels 1--3). fast$_1$ = correct and
faster than eager PyTorch. Geo-mean speedup is over correct samples only.
Gen./task is the maximum number of kernel-generator calls per task.}
\label{tab:headline}
\centering
\footnotesize
\setlength{\tabcolsep}{4.5pt}
\begin{tabular}{lccccc}
\toprule
Method & Compile\% & Correct\% & fast$_1$\% & Geo-mean & Gen./task \\
\midrule
Zero-shot   & 70 & 30 & 13 & 0.98$\times$ & 1 \\
\textbf{Guided} & \textbf{77} & \textbf{43} & \textbf{27} & 0.99$\times$ & 1 \\
Iterative   & 67 & 30 & 10 & 0.98$\times$ & $\le$3 \\
Agentic     & 73 & 30 & 10 & 0.97$\times$ & $\le$12 \\
\bottomrule
\end{tabular}
\end{table}

Profiler-guided prompting is the best single method (Table~\ref{tab:headline}).
With the same one generator call as zero-shot, it raises correctness from 30\% to
43\% and doubles fast$_1$ from 13\% to 27\%. Iterative refinement does not help
this model; retrying with the error log rarely turns a failure into a success.
The repaired agent matches zero-shot in aggregate while using up to 12$\times$ the
generator calls. But the aggregate hides a very different spread of successes.

\subsection{Difficulty decides the winner}

\begin{table}[t]
\caption{Correct solutions out of 10, per difficulty level.}
\label{tab:levels}
\centering
\small
\begin{tabular}{lccc}
\toprule
Method & L1 (single op) & L2 (fused seq.) & L3 (network) \\
\midrule
Zero-shot & 9 & 0 & 0 \\
Guided    & \textbf{10} & 0 & \textbf{3} \\
Iterative & 9 & 0 & 0 \\
Agentic   & 5 & \textbf{3} & 1 \\
\bottomrule
\end{tabular}
\end{table}

Table~\ref{tab:levels} is the main qualitative result. On L1 the simple methods
win (zero-shot and guided: 9 and 10 of 10) and the agent's machinery hurts
(5/10), because its analyzer talks the model into custom kernels where a one-line
torch op would pass. On L2 the agent is the only method that solves anything
(3/10 against 0/10 for the rest); naming the fusion up front and then checking it
by execution is what these tasks need. On L3 only guided prompting succeeds
(3/10), because the profile points the model at the layer that dominates runtime.
No method wins at every level, which matches what Snell et al.~\cite{snell2024}
report for test-time strategies on math problems.

\subsection{Component ablation and test-time compute}\label{sec:ablation}

\begin{table}[t]
\caption{Agentic component ablation. Every variant flips one component off against
the same best-of-4 backbone; $\Delta$ is correctness change (pp) vs.\ the full
system.}
\label{tab:ablation}
\centering
\small
\setlength{\tabcolsep}{5pt}
\begin{tabular}{lcccc}
\toprule
Variant & Correct\% & Compile\% & fast$_1$\% & $\Delta$ \\
\midrule
\textbf{Agentic (full, $n{=}4$)} & 30 & 73 & 10 & --- \\
$-$ RAG researcher       & 20 & 77 & 7  & $-10$ \\
$-$ Code analyzer        & 20 & 63 & 10 & $-10$ \\
$-$ Feedback analyzer    & \textbf{37} & 73 & 10 & $\mathbf{+7}$ \\
$-$ Refinement loop      & 20 & 70 & 10 & $-10$ \\
$-$ Best-of-$n$ ($n{=}1$) & 3  & 23 & 0  & $\mathbf{-27}$ \\
\bottomrule
\end{tabular}
\end{table}

\begin{figure}[t]
  \centering
  \includegraphics[width=\columnwidth]{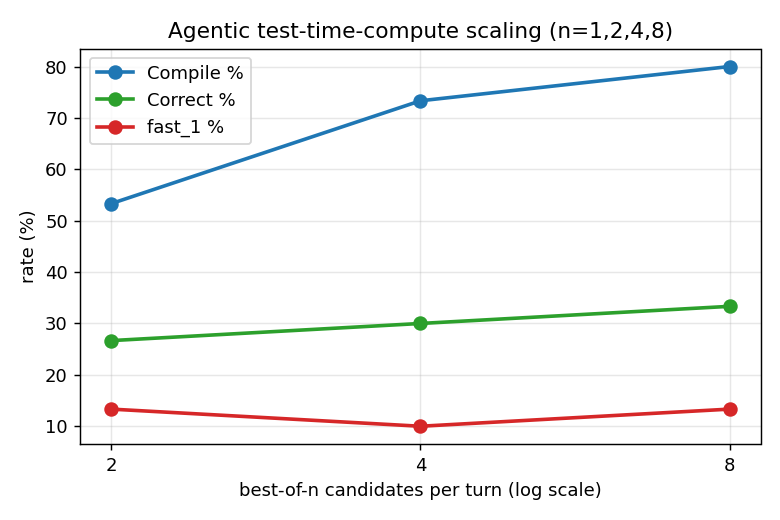}
  \caption{Test-time-compute scaling: full agent, varying only candidates per turn
  ($n\in\{1,2,4,8\}$; evaluator keeps the best). Correctness: 3/27/30/33\%;
  compile: 23/53/73/80\%. The first doubling makes the biggest difference;
  returns diminish after that.}
  \label{fig:bestof}
\end{figure}

Table~\ref{tab:ablation} and Fig.~\ref{fig:bestof} take the agent apart.
Best-of-$n$ sampling matters far more than anything else. Without it the system
drops to 3\% correct and 23\% compiling, no better than the unrepaired pilot on
correctness and worse on compile rate (23\% vs.\ 33\%). The analyzer, retrieval,
and the loop each add about 10 percentage points, but only on top of sampling.
The feedback analyzer slightly hurts: correctness rises 7 points when it is
removed. Its rewritten diagnoses sometimes over-specify the next attempt, for
example by telling the model to keep refactoring a design that already passes
(Appendix~\ref{app:output}), while the raw compiler error leaves the model free
to make a small fix. The scaling curve rises steeply and then flattens: going
from $n{=}1$ to $2$ adds 24 points of correctness, from $n{=}4$ to $8$ only 3
points for twice the evaluation cost.

\subsection{Error taxonomy}

\begin{table}[t]
\caption{Failure category counts over 30 tasks (correct excluded; ``slow'' =
correct but not faster than baseline). Categories are not exhaustive: one
zero-shot and one iterative failure fall outside them.}
\label{tab:taxonomy}
\centering
\small
\setlength{\tabcolsep}{5pt}
\begin{tabular}{lccccc}
\toprule
Method & compile & halluc.\ API & shape & wrong out & slow \\
\midrule
Zero-shot & 7 & 4 & 5 & 4 & 5 \\
Guided    & 5 & 6 & 3 & 3 & 5 \\
Iterative & 9 & 3 & 4 & 4 & 6 \\
Agentic   & 8 & 8 & \textbf{0} & 5 & 6 \\
\bottomrule
\end{tabular}
\end{table}

Table~\ref{tab:taxonomy} shows that the agent removes shape mismatches entirely,
so the analyzer's explicit reasoning about shapes and fusion works. It also
doubles hallucinated APIs, to 8 cases: pushed to write more custom code, the
model more often invents symbols, such as the non-public header
\texttt{ATen/native/cudnn/utils.h} or a call to \texttt{F.batch\_norm} from C++.
What remains is an API-grounding problem, not a planning problem.

\section{Discussion and Analysis}\label{sec:fail}

\subsection{Root cause 1: the cuBLAS/cuDNN ceiling (why ``slow'' happens)}

For every method, the geometric-mean speedup of correct kernels is
0.97--0.99$\times$: on average a correct kernel is a little slower than eager
PyTorch. The reason is structural. The L1 GEMM tasks multiply $4096^2$ matrices,
which is firmly compute-bound, and PyTorch hands them to cuBLAS kernels with
tiling and double-buffering tuned for the architecture. A plain
\texttt{load\_inline} kernel from a 7B model reaches a fraction of that
throughput, and its correct attempts only tie by calling \texttt{torch.matmul} in
the end. The speedups appear where the roofline model~\cite{roofline} says custom
code can win: memory-bound epilogues and reductions, the same regime in which
hand-written fused kernels such as FlashAttention~\cite{flashattention} pay off.
The agent's best speedup, 1.23$\times$, came on the skinny-$K$ GEMM
(\path{7_Matmul_with_small_K}), where cuBLAS is bound by launch overhead and
utilization. The guided method's speedups came from fusing bias, activation, and
clamp chains after convolutions, which cuts kernel launches and trips to global
memory. So the realistic goal for a small model is a correct fused kernel at
about 1$\times$, and methods differ mainly in correctness, which is what our
metrics show.

\subsection{Root cause 2: why every single-shot method scores 0/10 on level 2}

L2 tasks chain three to five operators. A passing \texttt{ModelNew} has to either
write one fused kernel, which is hard (the pilot showed the model cannot reliably
compile tiled multi-op CUDA), or keep the reference semantics exactly while
fusing only the easy parts, which is subtle:
\path{2_ConvTranspose2d_BiasAdd_...}, for instance, adds a learned bias whose
shape broadcasts differently from the convolution's own bias. Zero-shot, guided,
and iterative fail in one of two ways: ambitious fused CUDA that does not
compile, or torch-level rewrites that quietly change the semantics (the
wrong-output column of Table~\ref{tab:taxonomy}). The agent's analyzer writes
down the exact operator chain and its dataflow before generation, and the loop
gives a second chance after a semantic miss. Together that is worth 3/10 where
everything else gets 0. The successes are modest. The representative solution
(Appendix~\ref{app:output}) keeps \texttt{ConvTranspose2d} and \texttt{MaxPool}
as torch ops and fuses \texttt{clamp+mean} into one pass: a correct fusion at
1.01$\times$, not a hand-tuned kernel.

\subsection{Root cause 3: why the pilot agent lost to zero-shot}

The v1 failure was not bad luck. The scaffold asked for more than the model could
do: it pushed every task toward custom tiled CUDA, the model could not compile
it, each diagnosis asked for more optimization rather than less, and the
selection logic could throw away an earlier compiling kernel in favour of a later
broken one. Each repair in \S\ref{sec:iterate} addresses one of these. The
ablation confirms the diagnosis: with sampling removed ($n{=}1$), v2 falls back
to 3\%. The prompts alone do not fix a weak generator; many candidates checked by
a real verifier do.

That is the main lesson of this paper for agents built on small models for
verifiable tasks: spend compute on candidates that an oracle can check before
spending it on reasoning about any one candidate. Kevin~\cite{kevin} reports the
opposite ordering, with serial refinement scaling better than parallel sampling,
for a 32B model trained with multi-turn RL on this benchmark. The two results
agree if refinement only pays once the generator can act on execution feedback.
An RL-trained 32B model can. An untrained 7B model mostly cannot, as our
iterative baseline and the feedback-analyzer ablation show, so its extra turns
are better spent as extra samples.

\subsection{Root cause 4: infrastructure---generated kernels can hang the GPU}
\label{sec:infra}

Two production runs stalled mid-study. In the worst case, a sampled kernel
deadlocked at runtime (an infinite loop in device code), and because KernelBench
evaluates in-process and a Python signal cannot interrupt a running CUDA kernel,
that single candidate wedged a worker for 14 hours. We now evaluate every
candidate in a child process with a hard 240\,s wall-clock kill and score timeouts
as failures. Anyone who runs LLM-generated GPU code unattended needs the same.

\subsection{Limitations}

\textbf{Statistical power.} Thirty tasks, one model, one seed: differences of
1--2 problems are within noise, so we emphasize the large effects (the
3\%$\to$30\% repair, the 0-vs-3 L2 split, the best-of-$n$ curve) and not the
small gaps. The three non-sampling methods use greedy decoding and are close to
deterministic; the best-of-$n$ sweep is our only measure of sampling variance.

\textbf{Compute matching.} The agent uses up to $12\times$ the generator budget of
guided prompting and still trails it in aggregate (30\% vs.\ 43\% correct). We
have not run guided prompting with best-of-$n$ sampling, which would be the
natural compute-matched control and the obvious next experiment.

\textbf{Baseline and harness.} We compare against eager PyTorch, KernelBench's
default baseline, not \texttt{torch.compile}, which is stronger on fused
chains~\cite{kernelbench}. We use KernelBench's default correctness and timing
checks rather than the hardened protocol of robust-kbench~\cite{robustkbench}.
Since the speedups in this study are modest (geometric mean below 1$\times$ for
every method; the agent's best kernel reaches 1.23$\times$), exploitation of
harness loopholes is unlikely to have inflated our numbers, but we have not
audited for it.

\textbf{Scope.} Results are fp32 on Volta; on GPUs with tensor cores the
library-dominance ceiling may shift. Our claims about competence-matched
scaffolding are for small, untrained models only.

\section{Conclusion}

With a local 7B code model held fixed, how it is driven changes KernelBench
outcomes as much as any detail of the prompt, and more structure does not
reliably mean better results. Profiler-guided prompting does best overall (43\%
correct, 27\% faster than baseline) for one generator call. The multi-agent loop
is the only method that solves fused level-2 tasks, but it over-engineers simple
ones, and before its repairs it was worse than no agent. The repairs that mattered
were matching the agent's ambition to the model's ability and buying correctness
with best-of-$n$ sampling under a real evaluator: more test-time compute, not
deeper reasoning. Speed stays bounded by cuBLAS/cuDNN. What a small model can
realistically deliver is a correct fused memory-bound kernel at parity, and that
is where the methods differ.

Three things follow. Routing by difficulty (guided prompting on L1 and L3, the
agent on L2) would reach 53\% correct on this set with no new components.
API-grounded decoding would attack the hallucinated-API failures that now
dominate the agent's errors. And fine-tuning on the preference and reflection
trajectories the agent already logs is how Kevin~\cite{kevin} and
AutoTriton~\cite{autotriton} turned mid-size models into competitive kernel
writers.

\section*{Code and Data Availability}
The agent framework, experiment harness, and all raw artifacts (per-task prompts,
generations, evaluation logs, and the figures/tables in this paper) are available
at \url{https://github.com/pango0/KernelBench-Agentic}.

\section*{Acknowledgment}
Experiments were conducted on the Taiwania~2 GPU cluster.


\appendices

\section{Input Prompt (verbatim)}\label{app:prompt}

Base task prompt shared by all four methods, shown for level-1 problem~1
(\texttt{1\_Square\_matrix\_multiplication}); the architecture block changes per
task.

\begin{lstlisting}
You write custom CUDA operators to replace the pytorch operators in the given architecture to get speedups.

You have complete freedom to choose the set of operators you want to replace. You may make the decision to replace some operators with custom CUDA operators and leave others unchanged. You may replace multiple operators with custom implementations, consider operator fusion opportunities (combining multiple operators into a single kernel, for example, combining matmul+relu), or algorithmic changes (such as online softmax). You are only limited by your imagination.

You are given the following architecture:

import torch
import torch.nn as nn

class Model(nn.Module):
    """
    Simple model that performs a single square matrix multiplication (C = A * B)
    """
    def __init__(self):
        super(Model, self).__init__()

    def forward(self, A: torch.Tensor, B: torch.Tensor) -> torch.Tensor:
        return torch.matmul(A, B)

N = 2048 * 2

def get_inputs():
    A = torch.rand(N, N)
    B = torch.rand(N, N)
    return [A, B]

def get_init_inputs():
    return []

Note: The kernels should be optimized for FP32 (32-bit floating point) precision.

Optimize the architecture named Model with custom CUDA operators! Name your optimized output architecture ModelNew. Output the new code in codeblocks. Please generate real code, NOT pseudocode, make sure the code compiles and is fully functional. Just output the new model code, no other text, and NO testing code!
\end{lstlisting}

\noindent The \emph{guided} method appends a profiling block before the final
instruction (excerpt):

\begin{lstlisting}
Here is GPU profiling information for the reference PyTorch implementation (measured on the target GPU):

Total CUDA time (reference forward): <X> ms
Top operators by CUDA time (self time when available):
  1. <op>: <t> ms
  ...

Use this profiling data to guide your optimization: focus on the highest-cost operators, consider kernel fusion where multiple hot ops appear in sequence, and target memory-bound vs compute-bound behavior implied by the dominant kernels.

Important: PyTorch already dispatches matmul/linear and convolution to highly tuned cuBLAS/cuDNN kernels -- a hand-written CUDA replacement for a plain GEMM or conv will almost always be SLOWER, not faster. Do not reimplement those from scratch. Instead, keep using the torch operator for the heavy compute and win by (1) fusing the cheap memory-bound epilogue around it (bias + activation + scale + clamp into one pass) and (2) replacing pure elementwise / reduction / normalization chains, where a custom kernel can actually cut memory traffic. Correctness first: only optimize an operator if you can keep the result numerically identical to the reference.
\end{lstlisting}

\section{System Prompts (verbatim, final v2)}\label{app:sysprompts}

All four agent roles run on the same model.

\subsection{Code Analyzer}
\begin{lstlisting}
You are a senior GPU performance engineer. You analyse PyTorch reference modules and plan how to replace their operators with fast custom CUDA/HIP kernels. You are precise, concrete, and you never write the final kernel here.
\end{lstlisting}

\subsection{RAG Researcher}
\begin{lstlisting}
You are a research assistant for GPU kernel engineering. You are given retrieved documentation excerpts and must distil ONLY the techniques that are directly useful for the task. Ground every claim in the excerpts; never invent APIs. Cite the source tag (e.g. [memory_optimization.md#2]) after each point.
\end{lstlisting}

\subsection{Kernel Generator (the competence-matched v2 prompt)}\label{app:genprompt}
\begin{lstlisting}
You are an expert CUDA/HIP kernel engineer. Your FIRST priority is a ModelNew that compiles and matches the reference numerically; speed comes only after that. PyTorch already dispatches matmul/linear and convolution to highly tuned cuBLAS/cuDNN -- do NOT reimplement a plain GEMM or conv as a custom kernel, it will be slower. Keep the torch op for the heavy compute and add a custom kernel only where it clearly helps: fusing the memory-bound epilogue (bias/activation/scale/clamp) or replacing pure elementwise / reduction / normalization chains. When you do write a torch.utils.cpp_extension kernel, use load_inline(name=..., cpp_sources=..., cuda_sources=..., functions=[...]) -- note load_inline takes cpp_sources/cuda_sources, NOT a 'sources=' argument, and the pybind names in 'functions' must exactly match the C++ functions you define. Prefer one simple, correct fused kernel over several ambitious ones. You output ONLY a single ```python``` code block that defines class ModelNew(nn.Module) with the same interface as Model, plus get_inputs() and get_init_inputs(). No prose, no tests.
\end{lstlisting}

\noindent For comparison, the \textbf{v1 generator prompt} that produced the failed
pilot (3\% correct):
\begin{lstlisting}
You are an expert CUDA/HIP kernel engineer. You replace PyTorch operators with correct, high-performance custom kernels using torch.utils.cpp_extension (load_inline) or Triton. You output ONLY a single ```python``` code block that defines class ModelNew(nn.Module) with the same interface as Model, plus get_inputs() and get_init_inputs(). No prose, no tests.
\end{lstlisting}

\subsection{Feedback Analyzer}
\begin{lstlisting}
You are a debugging and performance triage expert for GPU kernels. Given an implementation and its evaluation result, you produce a crisp diagnosis and a concrete, prioritised fix list for the next attempt. You do NOT write the full kernel; you direct the fix.
\end{lstlisting}

\section{Representative Agent Output}\label{app:output}

\subsection{A correct level-2 solution (from the agent, the only method to solve any L2 task)}
Task \path{10_ConvTranspose2d_MaxPool_Hardtanh_Mean_Tanh}; final turn;
correct, 1.01$\times$. The shape of the solution is typical: torch ops are kept
for conv and pool, and \texttt{clamp+mean} is fused into one pass.

\begin{lstlisting}
import torch
import torch.nn as nn
from torch.utils.cpp_extension import load_inline

class ModelNew(nn.Module):
    def __init__(self, in_channels, out_channels, kernel_size, stride, padding,
                 maxpool_kernel_size, maxpool_stride, hardtanh_min, hardtanh_max):
        super(ModelNew, self).__init__()
        self.conv_transpose = nn.ConvTranspose2d(in_channels, out_channels,
            kernel_size, stride=stride, padding=padding)
        self.maxpool = nn.MaxPool2d(kernel_size=maxpool_kernel_size,
            stride=maxpool_stride)
        self.hardtanh = nn.Hardtanh(min_val=hardtanh_min, max_val=hardtanh_max)

    def forward(self, x):
        x = self.conv_transpose(x)
        x = self.maxpool(x)
        x = self.fused_hardtanh_mean(x)
        x = torch.tanh(x)
        return x

    @staticmethod
    def fused_hardtanh_mean(x):
        return torch.mean(torch.clamp(x, min=-1, max=1), dim=(2, 3), keepdim=True)

def get_inputs():
    return [torch.rand(batch_size, in_channels, height, width)]

def get_init_inputs():
    return [in_channels, out_channels, kernel_size, stride, padding,
            maxpool_kernel_size, maxpool_stride, hardtanh_min, hardtanh_max]
\end{lstlisting}

\subsection{A representative failure (hallucinated API, $-$RAG variant)}
Compiler output excerpt; the model invented a private ATen header and called a
Python API from C++:

\begin{lstlisting}
fuse_relu_bn/cuda.cu:7:10: fatal error: ATen/native/cudnn/utils.h: No such file or directory
    7 | #include <ATen/native/cudnn/utils.h>

main.cpp:11:22: error: 'F' was not declared in this scope
   11 |  auto bn_output = F.batch_norm(x, weight, bias, running_mean, running_var, true, eps, 0.0);
main.cpp:19:56: error: 'fuse_relu_bn_cuda' was not declared in this scope
   19 | m.def("fuse_relu_bn_cuda", torch::wrap_pybind_function(fuse_relu_bn_cuda), ...);
\end{lstlisting}

\subsection{A Feedback Analyzer diagnosis (illustrating its failure mode)}
Produced after turn~0 of the L2 task above, which was already correct. The
directive asks for more refactoring of a design that already works; this is the
behavior that makes correctness go up when the role is removed
(Table~\ref{tab:ablation}):

\begin{lstlisting}
CATEGORY: performance
ROOT_CAUSE: The primary bottleneck identified is the `fused_hardtanh_mean` function, which involves both clamping and mean reduction operations. These operations are memory-bound and can benefit from being fused with subsequent operations to reduce the overall computational overhead.
DIRECTIVE: Refactor the ModelNew class to use the combined `fused_hardtanh_mean_tanh` function instead of the separate `fused_hardtanh_mean` function, and verify if the performance improves. Additionally, investigate any potential optimizations for the Convolution Transpose and Max Pooling operations.
\end{lstlisting}

\section{Agent Modifications}\label{app:mods}

A chronological record of every modification and the failure that prompted it.

\textbf{C1. Generator/analyzer prompt inversion (``correct first'').}
v1 mandated custom \texttt{load\_inline}/Triton kernels for every task; v2
(Appendix~\ref{app:genprompt}) makes correctness the stated priority, forbids
reimplementing plain GEMM/conv, directs fusion to memory-bound epilogues, and pins
the exact \texttt{load\_inline} signature the model repeatedly hallucinated. The
analyzer's section instructions were rewritten the same way (flag matmul/conv as
``keep the torch op''). \emph{Effect (with C2): 3\%$\to$30\% correct,
33\%$\to$73\% compile.}

\textbf{C2. Best-of-$n$ sampling under the execution oracle.}
The generator samples $n{=}4$ candidates per turn (temperature 0.7); the
\emph{evaluator}---not the LLM---ranks them by
$(\textit{correct},\textit{compiled},\textit{speedup})$ and keeps the best.
The ablation (Table~\ref{tab:ablation}) shows this is the single most important
component ($-27$pp when removed); the sweep (Fig.~\ref{fig:bestof}) maps
$n\in\{1,2,4,8\}$.

\textbf{C3. Monotone final-kernel selection.}
Previously the submitted kernel could be the \emph{last} generated turn even if an
earlier turn compiled and the last did not. Selection now maximizes
$(\textit{correct},\textit{compiled},\textit{speedup},\textit{turn})$ over all
turns, so refinement can never make the submission worse. (The same fix was applied
to the iterative baseline, for fairness.)

\textbf{C4. Hard per-candidate evaluation timeout.}
Each candidate is compiled/checked/timed in a child process killed after 240\,s
(\path{agentic/evaluator.py:evaluate_kernel_safe}). Motivation: a generated
kernel deadlocked the GPU and---because a Python signal cannot interrupt a running
CUDA kernel---wedged a worker for 14\,h. Timeouts are scored as ordinary failures.

\textbf{C5. Optional reference fallback (kept OFF for all reported numbers).}
A flag can ship the unmodified reference (\texttt{ModelNew = Model}) when no turn
compiles---useful as a deployment floor, but disabled throughout this study because
a passthrough is trivially ``correct'' at 1$\times$ and would inflate benchmark
correctness.

\textbf{C6. Coherent ablation infrastructure.}
All agentic variants share one best-of-4 backbone configuration, so each ablation
flips exactly one component; the report generator renders the component table, the
best-of-$n$ sweep, and the scaling figure directly from raw evaluation JSON.


\begin{thebibliography}{00}

\bibitem{kernelbench} A.~Ouyang, S.~Guo, S.~Arora, A.~L.~Zhang, W.~Hu, C.~R\'e, and
A.~Mirhoseini, ``KernelBench: Can LLMs write efficient GPU kernels?,'' in
\emph{Proc. Int. Conf. Machine Learning (ICML)}, 2025, arXiv:2502.10517.

\bibitem{qwen25coder} B.~Hui \emph{et al.}, ``Qwen2.5-Coder technical report,''
arXiv:2409.12186, 2024.

\bibitem{pytorch} A.~Paszke \emph{et al.}, ``PyTorch: An imperative style,
high-performance deep learning library,'' in \emph{Advances in Neural Information
Processing Systems (NeurIPS)}, 2019.

\bibitem{cudnn} S.~Chetlur, C.~Woolley, P.~Vandermersch, J.~Cohen, J.~Tran,
B.~Catanzaro, and E.~Shelhamer, ``cuDNN: Efficient primitives for deep learning,''
arXiv:1410.0759, 2014.

\bibitem{roofline} S.~Williams, A.~Waterman, and D.~Patterson, ``Roofline: An
insightful visual performance model for multicore architectures,'' \emph{Commun.
ACM}, vol.~52, no.~4, pp.~65--76, 2009.

\bibitem{bm25} S.~Robertson and H.~Zaragoza, ``The probabilistic relevance
framework: BM25 and beyond,'' \emph{Found. Trends Inf. Retr.}, vol.~3, no.~4,
pp.~333--389, 2009.

\bibitem{aicudaengineer} R.~T.~Lange, A.~Prasad, Q.~Sun, M.~Faldor, Y.~Tang, and
D.~Ha, ``The AI CUDA Engineer: Agentic CUDA kernel discovery, optimization and
composition,'' Sakana AI, Tech. Rep., 2025.

\bibitem{robustkbench} R.~T.~Lange, Q.~Sun, A.~Prasad, M.~Faldor, Y.~Tang, and
D.~Ha, ``Towards robust agentic CUDA kernel benchmarking, verification, and
optimization,'' arXiv:2509.14279, 2025.

\bibitem{kevin} C.~Baronio, P.~Marsella, B.~Pan, S.~Guo, and S.~Alberti, ``Kevin:
Multi-turn RL for generating CUDA kernels,'' in \emph{Proc. Int. Conf. Learning
Representations (ICLR)}, 2026, arXiv:2507.11948.

\bibitem{cudal1} X.~Li, X.~Sun, A.~Wang, J.~Li, and C.~Shum, ``CUDA-L1: Improving
CUDA optimization via contrastive reinforcement learning,'' arXiv:2507.14111,
2025.

\bibitem{cudallm} W.~Chen, J.~Zhu, Q.~Fan, Y.~Ma, and A.~Zou, ``CUDA-LLM: LLMs can
write efficient CUDA kernels,'' arXiv:2506.09092, 2025.

\bibitem{astra} A.~Wei, T.~Sun, Y.~Seenichamy, H.~Song, A.~Ouyang, A.~Mirhoseini,
K.~Wang, and A.~Aiken, ``Astra: A multi-agent system for GPU kernel performance
optimization,'' arXiv:2509.07506, 2025.

\bibitem{tritonbench} J.~Li \emph{et al.}, ``TritonBench: Benchmarking large
language model capabilities for generating Triton operators,'' arXiv:2502.14752,
2025.

\bibitem{autotriton} S.~Li \emph{et al.}, ``AutoTriton: Automatic Triton
programming with reinforcement learning in LLMs,'' arXiv:2507.05687, 2025.

\bibitem{geak} J.~Wang \emph{et al.}, ``GEAK: Introducing Triton kernel AI agent
\& evaluation benchmarks,'' arXiv:2507.23194, 2025.

\bibitem{codex} M.~Chen \emph{et al.}, ``Evaluating large language models trained
on code,'' arXiv:2107.03374, 2021.

\bibitem{alphacode} Y.~Li \emph{et al.}, ``Competition-level code generation with
AlphaCode,'' \emph{Science}, vol.~378, no.~6624, pp.~1092--1097, 2022.

\bibitem{codet} B.~Chen, F.~Zhang, A.~Nguyen, D.~Zan, Z.~Lin, J.-G.~Lou, and
W.~Chen, ``CodeT: Code generation with generated tests,'' in \emph{Proc. Int. Conf.
Learning Representations (ICLR)}, 2023.

\bibitem{monkeys} B.~Brown \emph{et al.}, ``Large language monkeys: Scaling
inference compute with repeated sampling,'' arXiv:2407.21787, 2024.

\bibitem{snell2024} C.~Snell, J.~Lee, K.~Xu, and A.~Kumar, ``Scaling LLM test-time
compute optimally can be more effective than scaling model parameters,''
arXiv:2408.03314, 2024.

\bibitem{rag} P.~Lewis \emph{et al.}, ``Retrieval-augmented generation for
knowledge-intensive NLP tasks,'' in \emph{Advances in Neural Information
Processing Systems (NeurIPS)}, 2020.

\bibitem{selfrefine} A.~Madaan \emph{et al.}, ``Self-Refine: Iterative refinement
with self-feedback,'' in \emph{Advances in Neural Information Processing Systems
(NeurIPS)}, 2023.

\bibitem{selfdebug} X.~Chen, M.~Lin, N.~Sch\"arli, and D.~Zhou, ``Teaching large
language models to self-debug,'' in \emph{Proc. Int. Conf. Learning
Representations (ICLR)}, 2024.

\bibitem{reflexion} N.~Shinn, F.~Cassano, E.~Berman, A.~Gopinath, K.~Narasimhan,
and S.~Yao, ``Reflexion: Language agents with verbal reinforcement learning,'' in
\emph{Advances in Neural Information Processing Systems (NeurIPS)}, 2023.

\bibitem{metagpt} S.~Hong \emph{et al.}, ``MetaGPT: Meta programming for a
multi-agent collaborative framework,'' in \emph{Proc. Int. Conf. Learning
Representations (ICLR)}, 2024.

\bibitem{triton} P.~Tillet, H.~T.~Kung, and D.~Cox, ``Triton: An intermediate
language and compiler for tiled neural network computations,'' in \emph{Proc. ACM
SIGPLAN Int. Workshop on Machine Learning and Programming Languages (MAPL)}, 2019,
pp.~10--19.

\bibitem{tvm} T.~Chen \emph{et al.}, ``TVM: An automated end-to-end optimizing
compiler for deep learning,'' in \emph{Proc. USENIX Symp. Operating Systems Design
and Implementation (OSDI)}, 2018.

\bibitem{ansor} L.~Zheng \emph{et al.}, ``Ansor: Generating high-performance
tensor programs for deep learning,'' in \emph{Proc. USENIX Symp. Operating Systems
Design and Implementation (OSDI)}, 2020.

\bibitem{flashattention} T.~Dao, D.~Y.~Fu, S.~Ermon, A.~Rudra, and C.~R\'e,
``FlashAttention: Fast and memory-efficient exact attention with IO-awareness,''
in \emph{Advances in Neural Information Processing Systems (NeurIPS)}, 2022.

\end{thebibliography}
\end{document}